# Invariance Measures for Neural Networks

Facundo Manuel Quiroga[a,b], Jordina Torrents-Barrena[d], Laura Cristina Lanzarini[a], Domenec Puig-Valls[c]

[a]*Instituto de Investigación en Informática LIDI, Facultad de Informática, Universidad Nacional de La Plata, Argentina*
[b]*Becario Posdoctoral, Universidad Nacional de La Plata, Argentina*
[c]*Universitat Rovira i Virgili, Reus, España*
[d]*Large Format Printing, HP Inc., Sant Cugat, Spain*

**Abstract**

Invariances in neural networks are useful and necessary for many tasks. However, the representation of the invariance of most neural network models has not been characterized. We propose measures to quantify the invariance of neural networks in terms of their internal representation. The measures are efficient and interpretable, and can be applied to any neural network model. They are also more sensitive to invariance than previously defined measures. We validate the measures and their properties in the domain of affine transformations and the CIFAR10 and MNIST datasets, including their stability and interpretability. Using the measures, we perform a first analysis of CNN models and show that their internal invariance is remarkably stable to random weight initializations, but not to changes in dataset or transformation. We believe the measures will enable new avenues of research in invariance representation.

## 1. Introduction

Neural networks (NNs) are currently the state of the art for many problems in machine learning. In particular, convolutional neural networks (CNNs) achieve very good results in many computer vision applications [1].

However, NNs can have difficulties learning good representations when inputs are transformed. For example, the classification of texture or star images generally requires invariance to rotation, scale and/or translation transformation [2, 3], and face or body estimation models require pose-invariance [4, 5, 6].

The properties of invariance and equivariance explain how a model reacts to transformations of its inputs. Understanding the invariance and equivariance of a network [7, 8, 9, 10] or any system [11] can help to improve their performance and robustness. There are various ways to achieve invariance or equivariance in a model. However, how these properties are encoded internally in a trained NN is generally unknown.

In this work, we present simple, efficient and interpretable measures of invariance for NNs' *internal representations*. These measures allow understanding the distribution of invariance of NN's activations, and their structure after a suitable analysis.

The following subsections present and expand common definitions of invariance and Equivariance and summarize previous approaches for measuring invariance and equivariance.

*1.1. Invariance, Same-Equivariance and Equivariance Properties*

To deal gracefully with transformations, such as rotations of an image, requires the properties of *invariance* o *equivariance* in a network, with respect to a properly defined set of transformations $T = [t_1, t_2, ..., t_m]$ [1].

---

[1]In this work, we define T as a finite set of transformations. Using an infinite set is also possible but unnecessarily complicates other definitions. In case an approximately infinite set is desired, the number of transformations in the set can be increased as needed.

*Email address:* fquiroga@lidi.info.unlp.edu.ar (Facundo Manuel Quiroga)

A network f is invariant to a single transformation t if transforming the input x with t does not change the network output. Formally, $\forall x, f(t(x)) = f(x)$. The notion of invariance can be generalized such that if $T = [t_1 \ldots t_m]$ is a set of m transformations, then f is invariant to T whenever $\forall x, f(t_1(x)) = f(t_2(x)) = \cdots = f(t_m(x))$. As shown in figure 1 (a), an invariant function produces the same output for all transformations T of its inputs.

*Same-equivariance* is a property related to invariance. A function is same-equivariant if the *same* transformation t can be applied either to the input or the output of the function (figure 1 (b)). Therefore, f is same-equivariant to t if $f(t(x)) = t(f(x)) \; \forall x$ [2].

The generalization of both notions is *equivariance*. A function f is equivariant to t whenever f's output changes *predictably* if x is transformed by t. Formally, it is equivariant if there exists a corresponding function $t'$ such that $\forall$ x, we have $f(t(x)) = t'(f(x))$ [2]; the structure of $t'$ gives rise to the predictability. Note that in this definition, $t'$ acts on the activation $f(x)$, while t acts on the input x. Therefore the function $t'$ is not restricted to act in any way similar to t as is the case in same-equivariance. As an example, figure 1 (c) shows how rotations in the input can correspond to translations in the output.

The notions of equivariance and same-equivariance can be generalized to a set of transformations as with the invariance case.

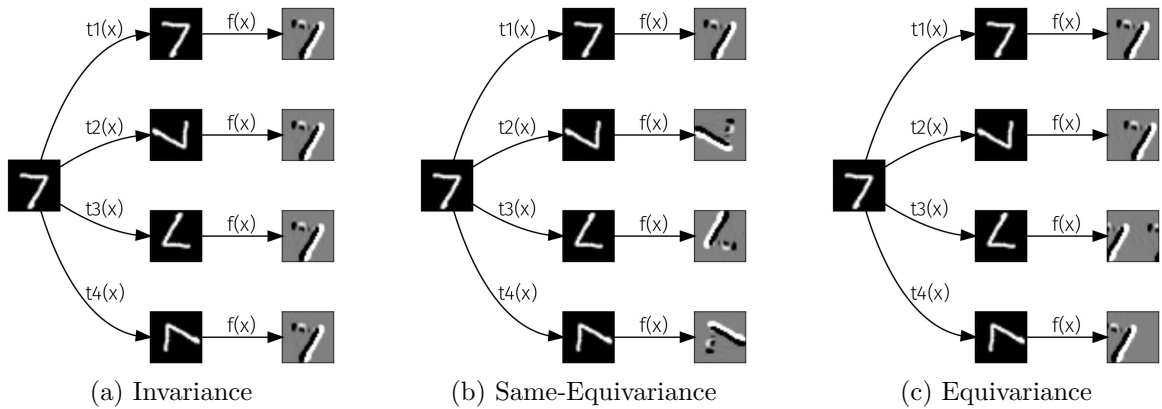

(a) Invariance  (b) Same-Equivariance  (c) Equivariance

Figure 1: Illustration of invariance, same-equivariance and equivariance of f(x) to the set of rotation transformations $T = [t_1, t_2, t_3, t_4]$ corresponding to rotations of $[0°, 90°, 180°, 270°]$ respectively and image inputs. In (a), the feature map calculated by f is the same regardless of the transformation of the input, and therefore f is invariant to T. In (b), the feature map is rotated in the same fashion as the input image and f is same-equivariant to T. In (c), rotations of the input correspond to translations of the feature map, where the magnitude of the translation is a function of the magnitude of the rotation angle, so that f is same equivariant to T, but the corresponding set $T' \neq T$ since it consists of translations.

*1.2. Invariance Measures*

Given a network and a set of transformations, invariance (and other related properties) can be measured in different ways.

To determine if f is equivariant to a transformation t that operates on inputs x, we need to find a corresponding transformation $t'$ that operates on the output space of f(x) [8]. A sufficient condition for the existence of $t'$ is that f is invertible. Since CNNs are approximately invertible[12, 8], the approximate existence of $t'$ is very likely. However, determining $t'$ can be very difficult; known approacher require assuming its functional form and estimating its parameters [8]. Therefore, empirically analyzing the equivariance of a CNN can be difficult [8].

Measuring invariance does not require an estimation of $t'$. Therefore, invariance can be computationally and conceptually easier to measure. Since invariance is a special case of equivariance, where $t'$ is the identity transformation, we can exploit this special structure to measure invariance in simple and efficient ways.

Most previous work has focused on empirically measuring the invariance of the *final output* of the network. The simplest invariance measures quantify the invariance of the final output of the network (ie. softmax layer for classification models) by measuring their changes with respect to a set of transformations of the input. Figure 2 (a) illustrates this pattern of analysis, where only the first and last layers of the network are taken into account.

Alternatively, we can consider a single node or layer, and measure its invariance with respect to its particular input and output, without taking into account the interactions with the rest of the network (figure 2 (b)). However, in general, these cases are simpler and can be handled analytically [13].

Given that most networks can be represented by an acyclic graph, we can generalize these notions by, for example, considering all intermediate values computed by the network as possible inputs or outputs. For simplicity, in the following, we will call these values *activations* of the network.

For example, we can consider all activations as outputs and measure their invariance with respect to the initial input to the network (figure 2 (c)). Alternatively, via a topological sort of the network graph, we can remove the first $k - 1$ nodes or layers from consideration, and calculate the invariance of the output of the network with respect to transformations of the input to node or layer k (figure 2 (d)). In summary, we can arbitrarily define a set of inputs to transform and a set of outputs to evaluate invariance or another such property. In this work, we focus on case (c), transforming only the input to the whole network and analyzing the impact of the transformation on all activations.

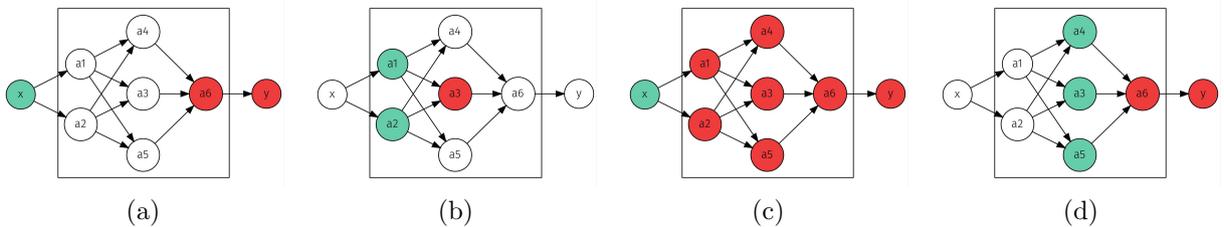

Figure 2: Different approaches for calculating invariance in a NN with activations $a_1, a_2, ... a_6$. Green nodes indicate transformed inputs, and red nodes indicate where the invariance is calculated. In (a), the invariance of the final output $y = a_6(x)$ of the network is measured with respect to transformations of the initial input x. In (b), the invariance of a single node or layer $a_3$ is calculated with respect to its direct inputs $a_1$ an $a_2$. In (c), the invariances of all nodes are calculated with respect to the input x. Finally, in (d) the invariance of the output y is calculated with respect to the output of activations $a_4$, $a_5$ and $a_6$.

*1.3. Contributions*

In this work, we focus on measuring the invariance of the *internal representation* of a network.

As explained before, the invariance of the activations of a network can be estimated by measuring how they change with respect to transformations of its inputs. In this work, we define measures that can empirically quantify the invariance of a NN with respect to a set of transformations following this principle. Since these invariance measures can quantify the invariance of activations or intermediate layers, they allow visualizing how invariant a network is as a whole and by layers or individual activations. Therefore, the measures can provide insights into how invariance is encoded and represented in a NN.

The measures can be applied to any NN, irrespective of its design or architecture, as well as any set of discrete and finite transformations. To keep the presentation short, in this work we focus on evaluating and applying these measures to CNNs, and affine transformations, since computer vision is an area where invariance is widely required and rotations, scaling, and translations are arguably the most common types of transformations.

In summary, our main contributions are:

1. A variance-based measure of the invariance of a NN, or indeed any model that computes an internal representation.
2. A distance-based alternative measure, which is a generalization of the variance case, as well as specializations for class-conditional invariances and convolutional layers.
3. An efficient method for computing the measures.

4. An evaluation of these measures and their variants, in terms of their validity and stability.
5. A comparison with the only other invariance measure described in the literature [7].

This work is organized as follows. Section 2 gives a summary of the state of the art for invariant and equivariant models, as well as previous measures related to the ones we propose. Section 3 describes the proposed measures. Section 4 presents experiments that form an empirical validation of the measures on a small but representative set of models, datasets, and transformations, along with their comparison and analysis. Finally, section 5 contains the conclusions and future work.

## 2. State of the art

In this section, we review previous measures proposed to quantify invariance. Invariance (and equivariance) can be measured indirectly, by observing the variations of the accuracy of a model as a function of the transformation of its inputs. In this way, several authors have defined several *accuracy-based* measures and diagrams. On the other hand, more detailed measures can be defined if the internal representation or *activations* of the network can be accessed; we will call these *activation-based* measures of invariance.

### 2.1. Accuracy-based measures

There have been various methods proposed to measure invariance in NNs indirectly by measuring the changes in the accuracy of the model with respect to changes in the input [14, 15, 16, 17, 18, 19, 20, 21, 22, 23]. In this case we refer to these as *accuracy-based* because these works mostly focus on measuring accuracy as a function of the transformation of the input but the same technique can be applied to other error functions such as the mean squared error.

In [22, 23, 24] the authors measure the effect of using different data augmentation schemes and CNNs architectures on the final accuracy of the method. To visualize the results, [23] proposed using a Translation Sensitivity Map that relates the classifier accuracy with the center position of the object in the image. In a similar fashion, [22, 23] used equivalent 1D plots to evaluate invariances to rotation and other transformations. [19, 16] focused on algorithms for determining the appropriate transformations to which the network is invariant, but also focused on accuracy as an indirect measure of invariance. Finally, invariance has been also studied from an adversarial perspective, since measuring invariance to transformations can be considered equivalent to measuring the effectiveness of an attack by considering the transformations as attacks[21, 20]. Therefore, measuring the required complexity of a transformation needed to decrease the accuracy of a classifier is equivalent to measuring the strength of an adversarial perturbation.

All of these invariance quantification methods focus on final accuracy instead of understanding the internal representations of the invariance in the network.

### 2.2. Activation-based measures

To the best of our knowledge, there are only two previously proposed measures of internal invariance or equivariance of the network [7, 8]. These have been used only once each [9, 25] after their publication to analyze specific properties of models. We briefly review these measures.

### 2.2.1. Invariance Measure of Goodfellow

Goodfellow et al. [7] were the first to propose a measure of invariance over the activations of the network instead of just the outputs. We recreate the definition of their measure here since it is conceptually similar to our own.

They define their measure in terms of the *firing rate* of activations. They assume that for each activation a there is an associated threshold U, so that if $a(x) > U$ then that unit is *firing*. Given a parametrized transformation $t(x, \gamma)$, where x is the input and $\gamma$ the parameter, the define a set of transformations of x as $T(x) = [t(x, \gamma) \mid \gamma \in \Gamma]$. parametrized by a finite set of parameters $\Gamma$. Their measure $GF(a, X)$ (equation 1) for activation a and dataset X is defined as the ratio between a *local* firing rate $Local(a, X)$ and a *global* firing rate $Global(a, X)$ (equations 2 and 3):

$$\text{GF}_U(a, X) = \frac{\text{Local}_U(a, X)}{\text{Global}_U(a, X)} \tag{1}$$

$$\text{Local}_U(a, X) = \frac{1}{|X|} \sum_{x \in X} \frac{1}{|T(x)|} \sum_{x' \in T(x)} f(x', U) \tag{2}$$

$$\text{Global}_U(a, X) = \mathop{\mathbb{E}}_{x \sim P(x)} [f(a, U)] \tag{3}$$

Where:

- $f(x, U) = \begin{cases} 1 & \text{if } a(x) > U \\ 0 & \text{otherwise} \end{cases}$.

- P(x) is a distribution over the samples.

While the measure is defined over an arbitrary set X, in practice the expectation in equation 3 is calculated over a fixed-size dataset of n samples. The threshold U, which can be different for different activations, is selected so that $G(a) = \alpha$, where $\alpha = 0.01$ in their experiments.

GF(a) is the invariance score for a single activation. To evaluate the invariance of an entire network N, they define Invp(N) as the mean of the top-scoring proportion p of activations in the deepest layer of N. They discard the $1 - p$ percentage of the least invariant activations on the hypothesis that *different sub-populations of units may be invariant to different transformations*. While this may be true for some datasets and models [10], they offer no method to determine the value of p and no support for that hypothesis. Indeed, different values of p may be required for different transformations. Finally, we note that this scheme is orthogonal to the definition of the measure since it can be applied to any invariance measure.

The measure has other difficulties as well for its practical application. First, the criteria for the selection of the threshold poses a performance problem since it requires calculating a percentile. The percentile cannot be computed in an online fashion unless using an approximation, and therefore requires storing in memory the n values of the activation a for the n samples and $\mathcal{O}(n\log(n))$ operations. Furthermore, there is no justification for the use $\alpha = 0.01$ to determine the value of U. Finally, the use of a threshold and the notion of the *firing rate* of an activation, while popular for models of biological NNs [26], has limited value for modern NN architectures.

Finally, since their work was presented in 2009, the measure was used to evaluate architectures that used layer-wise unsupervised pretraining and convolutional deep belief networks, with synthetic images and a custom video dataset. Since then, to the best of our knowledge only [9] have used their method to evaluate models for invariance in a limited fashion on the CIFAR10 and ImageNet datasets to then propose a new activation function.

*2.2.2. Equivariance measure of Lenc*

Lenc et al.[8] measured the *equivariance* of the activations of a CNN given by $y = f(x)$ with respect to a transformation T. The transformation was applied to the input x. In order to make possible the search for the equivariant function, they assume that the set of possible equivariances of the model f is a subset of the affine functions. That assumption allows optimizing the parameters of T′ after a network is trained via traditional gradient descent based algorithms using the error function $|L(T(x)) - T'(L(x))|$, where L is the function that computes a given convolutional layer with respect to the initial input to the network. A different transformation T′ was then estimated for each layer, using the total network error as a loss function. A particular distance [8] from $A_T$ to the identity matrix was utilized as an invariance measure of the layer's representation. Although this approach measures the equivariance, it *i)* only deals with affine transformations, which limits its applicability to convolutional layers as a spatial correspondence for the affine map is needed, *ii)* requires an optimization process, and *iii)* is not simple to interpret.

Since then, to the best of our knowledge only [25] used Lenc's measure to evaluate models and, in turn, modified the loss function to improve the equivariance and invariance capabilities of a model. However, the authors used the technique to estimate the impact of this new loss function only for the last network layer, and not the internal representation.

*2.3. Other related techniques*

*Measurement Invariance.* Also known as *Factorial Invariance* [27, 28], Measurement Invariance is a well-established field that seeks to provide statistical models with the property of measuring the same construct across different groups. For example, its techniques can be used to determine whether a certain measure is invariant to different race or gender groups, by analyzing the behavior of its latent variables, analytically or empirically. However, Measurement Invariance methods are focused on statistical modeling techniques such as Confirmatory Factor Analysis and cannot be applied directly to NNs [28].

*Ad-hoc invariance measures.* In many areas, models or quantities are have been hypothesized to be invariant to certain variables [11]. In turn, ad-hoc techniques have been created to test those hypotheses of invariances [11].

*Invariance measure of Quiroga et al..* Quiroga et al. [29] defined a variance-based invariance measure. This work extends that formulation by refining and expanding the definition of the measure, adding an alternative distance-based variance measure, and performing thorough validation of the measure. The details of the measures are described in Section 3.

*Qualitative Evaluation.* Another approach to determine the invariance or equivariance of features consists of evaluating them qualitatively. For example, visualizations have been used to understand invariance and other properties such as diversity and discrimination in CNNs [30, 31, 32, 10]. However, this approach is limited in that a useful visualization must exist for each type of feature and its interpretation can be subjective.

*Other Transformational Measures.* While invariance and equivariance are very important properties, other authors have defined measures for other properties such as feature complexity, invertibility, selectivity, capacity, and attention [33, 34, 35].

## 3. Measures

In this section, we define the Invariance measures along with a general framework for defining new transformational measures.

*3.1. General framework*

Our objective is to compute a measure of the activations of a model f with respect to a set of transformations T of its input x. In this case, our measures will be of invariance, but they could also target another type of property.

Given an input x, a neural network model contains computes many intermediate or hidden values, which we will call $a_1(x), ..., a_k(x)$. In the interest of brevity, we will call such values $a_i(x)$ *activations* of the network f. This term is not to be confused with *activation functions* such as ReLU or TanH. An activation *can* be the result of applying an activation function to a tensor, or simply the output of a convolutional or fully connected layer. Note that x always refers to the input of the whole network, and never to the input of an intermediate layer, as shown in Figure 3.

For example, let f be a two-layered network with a convolutional layer followed by a ReLU activation function and a fully connected layer. The output of a convolutional layer contains H × W × C scalar *activations*. After applying the ReLU, we obtain another set of H × W × C *activations*. By applying a flatten operation followed by a fully connected layer with D neurons to the output of the convolutional layer, we have another D *activations*. Therefore, there are k = H ∗ W ∗ C + H ∗ W ∗ C + D activations in this network.

In this case, we have ignored the output of the flatten operation. For appropriate cases such as this, a subset of activations can be ignored for the computation of the measure, since their output is simply a reshape of other activations. In other cases, outputs such as perhaps the activations of a convolutional layer before the ReLU is applied can also be ignored to reduce the amount of information to analyze.

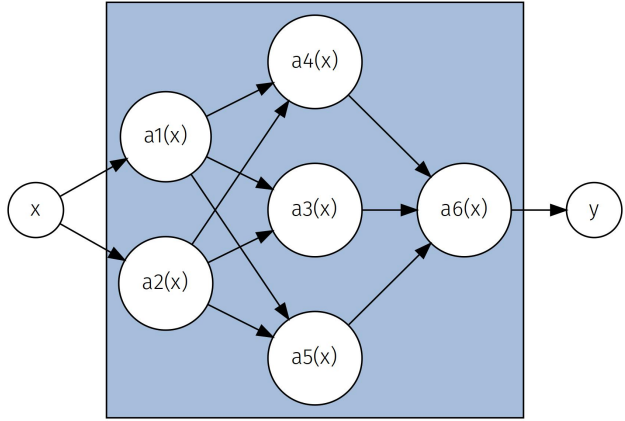

Figure 3: Diagram of a network f with its activations. Given an input x, the network computes its output y = f(x), by calculating its activations $a_1(x), \ldots, a_6(x)$. The final output value y is simply the value $a_6(x)$. Instead of considering each activation $a_i$ as a function of the output of other activations that feed into it, the activation is viewed as a function of the original input x.

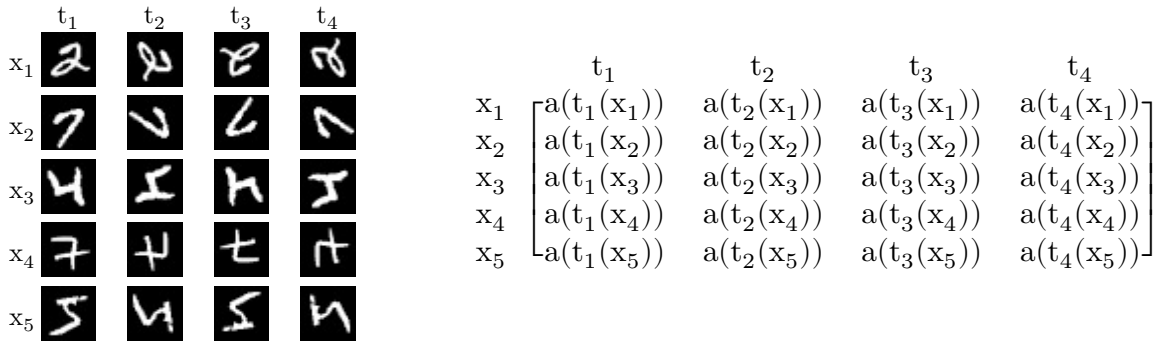

(a) Matrix of samples and transformations

(b) ST matrix of activations

Figure 4: (a) Matrix of samples and their corresponding transformations given by $t_j(x_i)$, for n = 5 samples and m = 4 transformations. (b) Corresponding ST matrix containing the activation values corresponding to each input for a single activation a.

To measure the invariance of the model f, we measure the invariance of the individual activations $a_1(x), \ldots a_k(x)$. Since the measure can be defined for an activation independently of the rest, we will focus on a single activation which we will denote simply as a(x).

### 3.2. Sample-Transformation activation matrix (ST)

In order to facilitate the definition of the measures, we define the concept of Sample-Transformations activation matrices (ST) (Figure 4), which provide the main context and notation for the transformational measures.

Given an activation a, a set of n samples $X = [x_1 \ldots x_n]$ and a set of m transformations $T = [t_1 \ldots t_m]$ defined over X, we can compute the value of a for all the possible transformations of the samples.

Let $x_{i,j} = t_j(x_i)$ be the sample obtained by applying transformation $t_j \in T$ to input $x_i \in X$. Given a, we define the sample-transformations activation matrix $ST(a, X, T)$ of size n × m as follows:

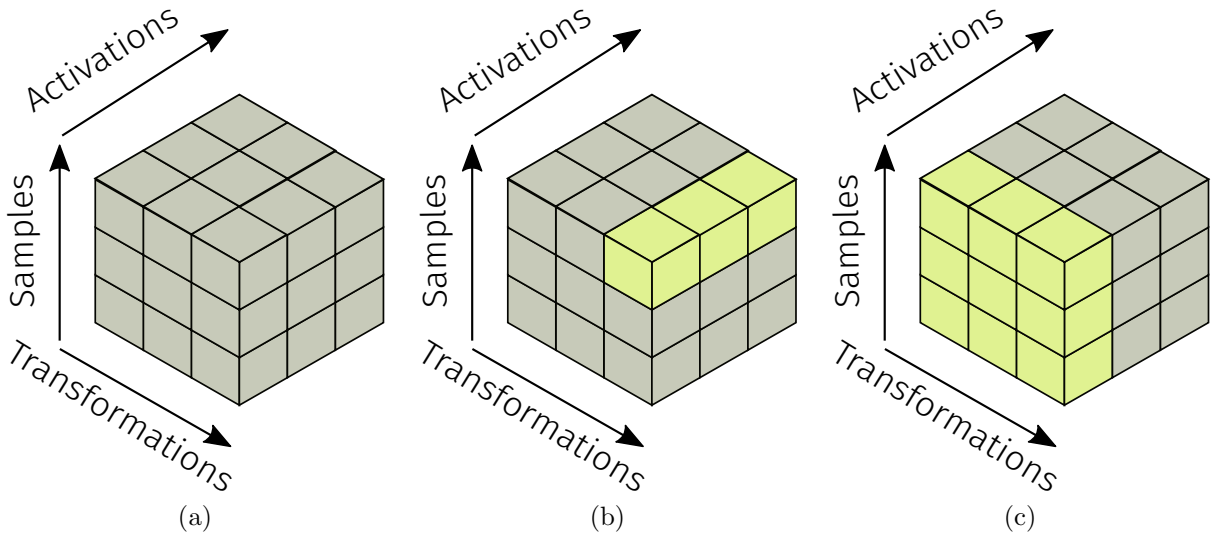

Figure 5: a) Cube of activation values for a network. b) The result of a forward pass for a single sample and transformation. c) Slice of the activations cube that corresponds to the ST a single output.

$$ST(a, X, T)_{i,j} = a(x_{i,j}) = a(t_j(x_i)) \qquad (4)$$

For simplicity, we will refer to $ST(a, X, T)$ as $ST(a)$ or simply ST whenever the context clearly determines a or X and T. Note that ST resembles the matrix of observations employed in a one-way ANOVA, where each transformation can be considered as a different *treatment*.

For a network f with k different activations, there are k associated ST matrices, which together form an $m \times n \times k$ cube (Figure 5).

Note that the result of a traditional forward pass only provides a small subset of the values of the ST matrix needed to compute a measure. Therefore the computation of the ST matrix must be performed in an online fashion for practical use to avoid storing all the activations for all the combinations of samples and transformations.

In the following subsections, we define three invariance measures based on the ST matrix, each based on different concepts. The ANOVA measure (section 3.3) uses the traditional analysis of variance procedure to determine if an activation is invariant or not assuming the various transformations are similar to treatments in an ANOVA setting. Variance-based measures [29] (section 3.4) use the common sample variance or standard deviation of each activation to quantify its invariance. Distance-based measures (section 3.6) compare the distance between activations to quantify how much they change under transformation of the inputs.

*3.3. ANOVA Measure*

The Analysis of Variance (ANOVA) is a statistical hypothesis testing method [36]. It is used to analyze samples of different groups. ANOVA can establish if the means for different groups of samples, called treatments, are statistically similar. While ANOVA is a parametric method, it has mild assumptions and is robust to violations of normality, especially with large sample sizes such as those available for machine learning datasets [36].

One-way ANOVA employs a matrix of observations that contain n rows and m columns, where each row corresponds to a sample, and m observations correspond to treatments that have been applied independently. The null hypothesis in ANOVA states that the means are the same for the different treatments/columns. Hence, we can adapt the interpretation of the method for invariance testing. The ST matrix can be interpreted as the matrix of observations in one-way ANOVA, where each treatment is a different transformation.

Therefore, the null hypothesis is equivalent to invariance, since transformations do not affect the activation. On the other hand, if the null hypothesis is rejected for an activation, then it is not invariant.

We define the **ANOVA** measure (AM) simply as the application of the one-way ANOVA procedure to the ST matrix of each activation, independently. Therefore, the only parameter of the measure is the significance value of the test, $\alpha$.

While the ANOVA tests are independent for each activation, the number of activations in a NN is quite large. Therefore, in the cases where the invariance of various activations is evaluated in tandem, we must apply a Bonferroni correction [36] to $\alpha$ to account for the large number of corresponding hypothesis tests. Finally, in order to be consistent with the rest of the measures, when the ANOVA measure detects a variant activation (rejection of null hypothesis) it outputs 1 as a result and 0 otherwise.

The choice of invariance as the null hypothesis can be considered strange since in general it is a property models are not assumed to have a priori. However, in many cases, the models that are being measured have been trained or designed for invariance. Therefore, while both approaches have their merits, using invariance as the null hypothesis can be more appropriate in many cases.

The computation of the ANOVA requires two iterations over the ST matrix. The first iteration computes the means for each treatment/transformation; the second, the within-group sum of squares. Both iterations are performed iterating first over transformations and then over samples, that is, iterating over the rows of the ST matrix.

*3.4. Variance-based Invariance Measures*

Variance-based invariance measures use the variance of an activation as an approximate notion of invariance. The variance Var is a function with range $[0, \infty)$. Therefore, we can consider an activation as exactly invariant to a set of transformations T if its variance is 0 with respect to the input x after being transformed by elements of T. Values greater than 0 indicate different degrees of lack of invariance. Since the variance with respect to the transformations depends on the scale of the activations, which can vary between activations, layers, and datasets, we measure two different sources of variance, Transformation Variance and Sample Variance, each computed by varying the transformations and samples, respectively. Afterward, these two variances are combined to obtain a normalized Normalized Variance measure which is dimensionless and can be used to compare the invariance of different activations. The following section describes the three measures and their relationship.

*3.4.1. Transformation Variance Measure*

The **Transformation Variance** (TV) of an activation a is defined as the mean variance of an activation, where the mean is computed over the set of samples and the variance over the set of transformations. This is equivalent to computing the average variance of each row in the ST matrix (equation [5]).

$$\mathrm{TV} = \mathrm{Mean}\left(\begin{bmatrix} \mathrm{Var}(\mathrm{ST}[1,:]) \\ \cdots \\ \mathrm{Var}(\mathrm{ST}[n,:]) \end{bmatrix}\right) \quad (5)$$

Where:

- $\mathrm{ST}[i,:] = \begin{bmatrix} \mathrm{ST}[i,1] & \cdots & \mathrm{ST}[i,m] \end{bmatrix}$ is a vector containing row i of ST(a).

- $\mathrm{Var}(\begin{bmatrix} x_1 & \cdots & x_n \end{bmatrix}) = \frac{\sum_{i=1}^{n} x_i - \bar{x}}{n-1}$ is the standard sample variance defined over a vector of observations $\begin{bmatrix} x_1 & \cdots & x_n \end{bmatrix}$.

- $\bar{x} = \mathrm{Mean}(\begin{bmatrix} x_1 & \cdots & x_n \end{bmatrix}) = \frac{\sum_{i=1}^{n} x_i}{n}$ is the standard sample mean.

Each row i of the ST matrix contains the activations for sample $x_i$ and all transformations (equation [4] and figure 4). Therefore, the variance is computed over the activations for different transformations; and the mean over samples. Should the activation be completely invariant to T, all the values in each row would

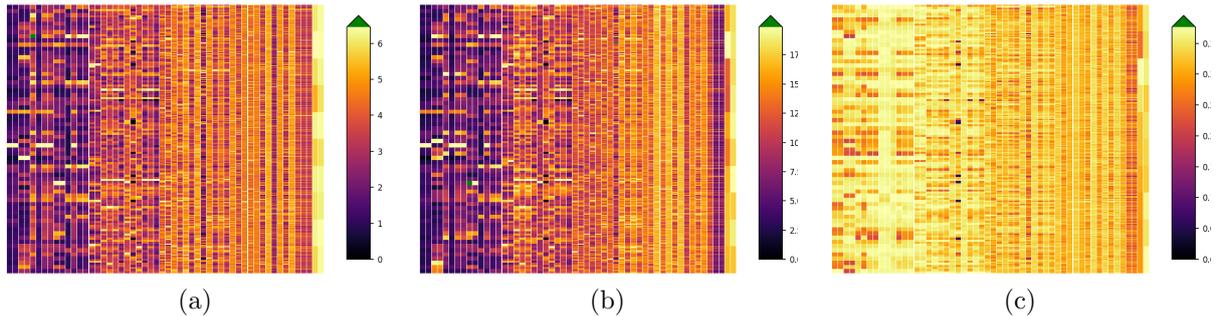

Figure 6: Calculation of the Transformation Variance measure for n = 5 samples and m = 4 transformations. First, (1) the variance of each row (over transformations) is calculated; then, (2) the mean of each column (over samples).

(a) (b) (c)

Figure 7: (a) Transformation Variance, (b) Sample Variance and (c) Normalized Variance of each activation of the ResNet model. The transformation set for this example is a set of 16 rotations distributed uniformly between 0° and 360°, and the dataset is the test set of CIFAR10.

be equal and therefore the variance of each row would be 0. In this way, if an activation is invariant to transformations then its *transformational* variance is 0.

Figure 7 (a) shows a visualization of the values of the Transformation Variance for all activations of the network in terms of a specialized heatmap. The heatmap is organized first by columns, where each column corresponds to a layer. Within each layer/column, there is a cell for each activation of the layer. Cell colors correspond to different levels of invariance. Values in green indicate outliers. Each layer/column can have a different number of activations and therefore rows. Since the order of activations within each layer is arbitrary, there is no row-structure in the image. Only layers of activation functions have the same structure as the previous layer because they act element-wise.

For layers that output a set of feature maps (convolutional, max-pooling, etc), we calculate the mean invariance over the spatial dimensions of each feature map. Therefore, if a layer's activation has size H × W × C, this tensor is collapsed into a C size vector (see section 3.5 for more details).

Using Welford's running mean and variance equations [37], computing the Transformation Variance requires a single iteration through the rows of ST, and therefore the running time is $\mathcal{O}(k \times n \times m)$.

The Transformation Variance is measured in units of the activation a. When TV(a) = 0 the activation is invariant to transformations. However, if TV(a) > 0 there is no clear interpretation for that value since the unit of the activation depends on both the samples employed and the parameters of the model, which may vary overmuch as can be observed in Figure 7 (a). To neutralize this unwanted source of variability, we can normalize the Transformation Variance values. We propose using the Sample Variance, as defined below, to divide the Transformation Variance and obtain a Normalized Variance measure.

*3.4.2. Sample Variance Measure*

The **Sample Variance** (SV) is the conceptual transpose of the Transformation Variance. It is equivalent to computing the Transformation Variance on the transpose of the ST matrix. Therefore, instead of computing the variance of each row(variance over transformations) and then computing the mean of the result, the Sample Variance is obtained by first computing the variance of each column (variance over samples with the same transformation), and then the mean over the resulting row vector (8).

$$
\overset{\rightarrow\ (2)\ \text{Mean}}{\underset{(1)\ \text{Var}\ \downarrow}{\begin{bmatrix} a(t_1(x_1)) & a(t_2(x_1)) & a(t_3(x_1)) & a(t_4(x_1)) \\ a(t_1(x_2)) & a(t_2(x_2)) & a(t_3(x_2)) & a(t_4(x_2)) \\ a(t_1(x_3)) & a(t_2(x_3)) & a(t_3(x_3)) & a(t_4(x_3)) \\ a(t_1(x_4)) & a(t_2(x_4)) & a(t_3(x_4)) & a(t_4(x_4)) \\ a(t_1(x_5)) & a(t_2(x_5)) & a(t_3(x_5)) & a(t_4(x_5)) \end{bmatrix}}} \implies \text{Mean}\left( \left[ \text{Var}\left( \begin{bmatrix} a(t_1(x_1)) \\ a(t_1(x_2)) \\ a(t_1(x_3)) \\ a(t_1(x_4)) \\ a(t_1(x_5)) \end{bmatrix} \right), \text{Var}\left( \begin{bmatrix} a(t_2(x_1)) \\ a(t_2(x_2)) \\ a(t_2(x_3)) \\ a(t_2(x_4)) \\ a(t_2(x_5)) \end{bmatrix} \right), \text{Var}\left( \begin{bmatrix} a(t_3(x_1)) \\ a(t_3(x_2)) \\ a(t_3(x_3)) \\ a(t_3(x_4)) \\ a(t_3(x_5)) \end{bmatrix} \right), \text{Var}\left( \begin{bmatrix} a(t_4(x_1)) \\ a(t_4(x_2)) \\ a(t_4(x_3)) \\ a(t_4(x_4)) \\ a(t_4(x_5)) \end{bmatrix} \right) \right] \right)
$$

Figure 8: Calculation of the Sample Variance measure for n = 5 samples and m = 4 transformations. First, the variance of each column (over samples) is calculated; then, the mean of each row (over transformations).

Equation 6 shows the formal definition of the Transformation Variance for an activation a in terms of its ST matrix.

$$SV = \text{Mean}\left( [\text{Var}(ST[:,1]) \quad \cdots \quad \text{Var}(ST[:,m])] \right) \tag{6}$$

While the Transformation Variance measures the variance due to the transformations of the samples, the Sample Variance measures the variance due to the natural variability of the domain. Figure 7 (b) shows the results of calculating the Sample Variance as a heatmap. Note that the order of magnitude of the values of the Sample Variance is similar to that of the Transformation Variance, and also depends on the layer and activation.

Using running mean and variance equations, computing the Sample Variance requires a single iteration through the columns of ST, and therefore the running time is also $\mathcal{O}(k \times n \times m)$.

*3.4.3. Normalized Variance Measure*

The **Normalized Variance** (NV) is simply the ratio between the Transformation Variance (equation 5) and the Sample Variance (equation 6) [2]:

$$\text{NV}(a) = \frac{\text{TV}(a)}{\text{SV}(a)} = \frac{\frac{1}{n}\sum_{i=1}^{n}\text{Var}(ST(a)[i,:])}{\frac{1}{m}\sum_{i=1}^{m}\text{Var}(ST(a)[:,i])} \tag{7}$$

The Normalized Variance is therefore a ratio that weights the variability due to the transformation with the sample variability. Since both have the same unit, the result is a dimensionless value. Figure 7 (c) shows the result of the Normalized Variance measure.

The computation of NV requires only two loops over the ST matrix, a transformation-first loop to compute the Transformation Variance, and a samples-first loop to compute the Sample Variance. Therefore, its running time is also $\mathcal{O}(k \times n \times m)$.

*Interpretation of values of the Normalized Variance.* We can analyse the possible values of NV as follows:

- If $\text{NV}(a) = 0$, then $\text{TV}(a) = 0$ and the activation is clearly invariant.

- If $\text{NV}(a) < 1$, the variance due to the transformations is less than that due to the samples, and so we can consider the activation to be approximately invariant.

- If $\text{NV}(a) > 1$ the same reasoning applies, but with the opposite conclusion.

- If $\text{NV}(a) = 1$, then both variances are in equilibrium, and there's no distinction between sample and transformation variability. In this case, it is possible that the dataset/domain naturally contains transformed samples, or simply that the model was trained in such a way that these values are similar.

Note that we can only interpret NV(a) in terms of relative invariance. For example, if $\text{NV}(a) = 0.5$, then the sample variance is twice the transformation variance.

---

[2]Note that the Normalized Variance measure corresponds to the V measure previously described by the authors in [29]

*Special cases in the computation of the Normalized Variance measure.* In cases where both TV(a) = 0 and SV(a) = 0, we have the very definition of dead activations which do not respond to any pattern. Therefore these have no use for the network, and so we set NV = 1 as a compromise. Alternatively, when SV(a) = 0 but TV(a) > 0 we set NV(a) = $+\infty$. This case is similar to the previous one but the transformations do make the activation vary in this case, possibly because they generate samples outside of the original distribution, and therefore indicate anomalous behavior in the activation.

Both cases can occur in common datasets, especially if they are synthetic or have been heavily preprocessed. For example, if all images in a dataset contain a black background and centered objects, it is likely that activations corresponding to the borders of feature maps, especially in the first layers, are trained to expect black pixels. However, transformations such as scale or translation can cause a shift in distribution so that the borders of some transformed images now contain non-black pixels. Hence the feature maps will possibly have non-zero activations in their edges so that SV(a) $\sim$ 0 but TV(a) > 0.

*Alternative definitions of* NV. The definition of NV may seem arbitrary at first. Indeed, it would be possible, for example, to further transform the result of the measure, for example using a logistic function to squash the measure to the interval [0, 1), so that NV(a) = 0.5 indicates equilibrium between TV and SV. We deliberately choose not to do this in order to preserve the notion that we lack a proper theory of approximate invariance to interpret these values.

Also, we preserve the definition in which NV(a) = 0 indicates invariance, since it reinforces the notion that we are calculating the *variance* of the activations.

We considered other alternatives such as NV(a) = logistic(TV(a) − SV(a)), but prefer the definition in which NV can be interpreted as a ratio. Since the variance or standard deviation depends on the scale of the activations, using the difference TV(a) − SV(a) would still tie the value of the measure to the scale. The coefficient of variation $\frac{\sigma}{\mu}$ could also be a viable alternative, but it would pose difficulties for many networks that are designed so that $\mu = 0$ for their activations.

Finally, for numerical stability reasons, we can replace the variance function for the standard deviation for the actual computation of the measure with only a slight overhead and no difference in its interpretation.

*3.5. Measure specialization for Feature maps*

Some types of layers can require specialization of the measures to obtain more useful results. Convolutional layers currently provide state of the art performance for several types of data including images. We describe a specialization of the variance measures for 2D convolutional layers since these are typically used with images; generalizing to 1D or ND convolutions is simple from this particular case.

Typical 2D convolutional layers output $K_f$ feature maps, each of size H × W. Therefore, the number of individual activations is $K_f$ × H × W, which can be considerably large for inputs with high spatial resolution. More importantly, the activations of a feature map have a spatial structure, which if ignored can yield uninteresting or incorrect results. For example, for object classification, the borders of the feature map usually yield little information, and analyzing them individually for invariance can be misleading or uninteresting.

Alternatively, we can measure the variance of feature maps by first aggregating the variance over the spatial dimensions. Given feature map F of size H × W such that F(i, j) is the activation in the i, j spatial coordinates, we can define TV(F) and SV(F) as (see Equation 8):

$$\begin{aligned} \text{TV(F)} &= \frac{1}{\text{H} \times \text{W}} \sum_{i=1}^{\text{H}} \sum_{j=1}^{\text{W}} \text{TV(F(i,j))} = \text{E} \\ \text{SV(F)} &= \frac{1}{\text{H} \times \text{W}} \sum_{i=1}^{\text{H}} \sum_{j=1}^{\text{W}} \text{SV(F(i,j))} \\ \text{NV(F)} &= \frac{\text{TV(F)}}{\text{SV(F)}} \end{aligned} \qquad (8)$$

Note that in the case of the NV measure, the aggregation is done at the level of the TV and SV measures, before the normalization. The alternative definition $\text{NV}_{\text{after}}$ (equation 9) is also possible but introduces potential problems, since individual ratios $\frac{\text{TV}(F(i,j))}{\text{SV}(F(i,j))}$ can vary wildly and produce less consistent values.

$$\text{NV}_{\text{after}}(F) = \frac{1}{H \times W} \sum_{i=1}^{H} \sum_{i=1}^{H} \sum_{j=1}^{W} \text{NV}(F(i,j)) \tag{9}$$

We chose to aggregate the variances of the feature map via the mean of the measures of each activation in the feature maps, so that TV(F) and SV(F) represent its mean variance. Alternatives for the aggregation could include max, min, sum or other such functions. Since feature maps are generally sparse[38], and given that filters may be active only in certain spatial regions, aggregating activations via the min of the activations instead of the mean would significantly underestimate the variance of the feature map; choosing max may cause the inverse problem. Note that for the NV measure, the sum would yield the same result as the mean (equation [8]), but not for TV or SV; in those cases, using the mean might be useful to disentangle the value of the measure from the size of the feature map.

### 3.6. Distance-based Invariance Measures

The previously shown variance-based measures use the variance as an indicator of invariance, quantifying deviations from the mean. This implies the assumption that the distribution of activations is roughly unimodal. However, in some cases, that assumption may be incorrect. For example, when the samples are drawn from different classes of objects, some activations may be triggered only by some subset of the classes.

Distance-based measures are similar to variance methods, but instead of calculating the variance, they employ a distance function between activations for different transformations or samples. By computing the distance between all pairs of activations, there are no assumptions of unimodality, and an appropriate distance function can be employed for different types of activations.

In the same way as with the variance measures, we define the TransformationDistance (TD), SampleDistance (SD) and NormalizedDistance (ND) measures using distance functions as follows (Equation 10)

$$\begin{aligned} \text{TD}(a) &= \text{Mean}\left(\begin{bmatrix} D(ST[1,:]) & \cdots & D(ST[n,:]) \end{bmatrix}\right) \\ \text{SD}(a) &= \text{Mean}\left(\begin{bmatrix} D(ST[:,1]) & \cdots & D(ST[:,m]) \end{bmatrix}\right) \\ \text{ND}(a) &= \frac{\text{TD}(a)}{\text{SD}(a)} \end{aligned} \tag{10}$$

Where D computes the mean distances between all values of a vector. Given an arbitrary distance function $d : R^2 \to R$, D is defined as:

$$D\left(\begin{bmatrix} x_1 & \cdots & x_n \end{bmatrix}\right) = \frac{\sum_{i=1}^{n} \sum_{j=1}^{n} d(x_i, x_j)}{n^2} \tag{11}$$

The distance measures calculate the average pairwise distances between activations either row-wise (Transformation Distance) or column-wise (Sample Distance), analogously to the Transformation Variance and Sample Variance measures. If an activation is completely invariant to a transformation, the mean distance between the activations for all transformations of a sample (D(ST[i, :])) will be 0. Therefore the Transformation Distance will be 0, and so will the Normalized Distance.

If the activation is approximately invariant, the mean distance between transformed samples will quantify this, and the mean distance between samples will quantify the degree of invariance.

| #Activations | Batch Size | Activations (MB) |
|---|---|---|
| 1M | 8 | 32 |
| 1M | 64 | 256 |
| 1M | 512 | 2048 |
| 10M | 8 | 320 |
| 10M | 64 | 2560 |
| 10M | 512 | 20480 |

Figure 9: Maximum RAM usage for the calculation of the Normalized Variance and Normalized Distance, assuming *float32* precision for activations.

*Approximation of the average distance.* The full computation of all distances between transformations and samples can be prohibitive. Such distance matrix would have size n × n for the sample variance and m × m for the transformation variance. While the mean distance does not require storing all distance values, it does require storing all k activations. As discussed before, the computation of the ST matrix must be done online. Therefore, we must employ an approximation to compute the mean distances. Since looping over the ST matrix is done by batches, it is straightforward to only compute distances for samples in the same batch. In this way, we approximate the mean of the full distance matrix by only computing the mean of the distances between blocks of the distance matrix.

A more principled approach would involve computing a low-rank approximation of the full euclidean distance matrix and then computing the mean distances [39]. However, the current best randomized algorithms for a single matrix are $\mathcal{O}(n+m)$ [40], which would render the computation of the measure impractical given a large number k of activations.

### 3.6.1. Running time

Analogously to the Normalized Variance case, the computation of the Normalized Distance requires two iterations over the ST matrix, one for the Transformation Distance and another for the Sample Distance. Given that in this case we are computing distances between all elements of a batch, the algorithm is $\mathcal{O}(b^2 \times \frac{n \times m}{b} \times k)$, where b is the batch size and k the number of activations.

Distance-based measures have an additional runtime factor b, compared with variance-based measures. The greater the value of b, the better the approximation, but also the running time and storage, since the computation of the mean pairwise distance is $O(b^2)$. Figure 9 shows examples of ram usage for typical networks.

The batch size b is strongly limited by the amount of RAM memory required to store activations, whether in CPU or GPU. Therefore, the approximation is, in principle, limited by the storage capacity. Nevertheless, several passes can be made to improve the approximation if required. By changing the order of iteration of the samples or transformations and aggregating the results of each pass, we can effectively sample more values of the distance matrix. Consequently, distance-based measures can choose the batch size independently from the desired number of samples for the approximation.

*Relationship between variance and squared euclidean distance.* In the case of the squared euclidean distance measure, it is well known that the variance and distance computations are equivalent, and so are the measures, as per equation [12]. In this particular case, we can avoid the approximation introduced by sparsely sampling the distance matrix by simply computing variance-based measures.

$$D([\,x_1\ \cdots\ x_n\,]) = \frac{\sum_{i=1}^{n}\sum_{j=1}^{n}(x_i - x_j)^2}{n^2}$$
$$= 2n\frac{\sum_{i=1}^{n}(x_i - \text{Mean}([\,x_1\ \cdots\ x_n\,]))^2}{n^2}$$
$$= 2\frac{\sum_{i=1}^{n}(x_i - \text{Mean}([\,x_1\ \cdots\ x_n\,]))^2}{n} \quad (12)$$
$$= 2\,\text{Var}([\,x_1\ \cdots\ x_n\,])$$

*Feature maps.* As mentioned before, measuring the invariance of each activation of a feature map may be undesirable, since the feature map has a spatial structure. The method described for the Normalized Variance measure to calculate the variance of feature maps (Section 3.5) is also valid for the distance measure (equation 13). That is, we can calculate the distance measure of each individual activation ND(F(i,j)) in a feature map F and then sum the individual measures.

$$\text{TD}(F) = \frac{1}{H \times W}\sum_{i=1}^{H}\sum_{j=1}^{W}\text{TD}(F(i,j))$$
$$\text{SD}(F) = \frac{1}{H \times W}\sum_{i=1}^{H}\sum_{j=1}^{W}\text{SD}(F(i,j)) \quad (13)$$
$$\text{ND}(F) = \frac{\text{TD}(F)}{\text{SD}(F)}$$

The advantage of a distance-based measure, however, is that we can use specialized distances for each type of activation or layer. Note that these distances can now be not between individual activations of a feature map, but between entire feature maps. Therefore, we may employ any image-based distance measure. For example, we can compare entire feature maps with a semantic distance measure such as the Frechet-Inception distance [41] and obtain pairwise distances between them.

Let F be a feature map, and ST(F) be a n × m matrix as before, but now each element $ST(F)[i,j] \in R^{h\times w}$ consists of the feature map calculated from sample i after applying transformation j. Then we can define TD, SD and ND in a similar fashion as equation [10], but for feature maps F (equation [14]).

$$\text{TD}(F) = \text{Mean}([\,D(ST(F)[1,:])\ \cdots\ D(ST(F)[n,:])\,])$$
$$\text{SD}(F) = \text{Mean}([\,D(ST(F)[:,1])\ \cdots\ D(ST(F)[:,m])\,]) \quad (14)$$
$$\text{ND}(F) = \frac{\text{TD}(a)}{\text{SD}(a)}$$

Where now d is a distance function between feature maps ($R^{h\times w}$) instead of real numbers:

$$D([\,F_1\ \cdots\ F_n\,]) = \frac{\sum_{i=1}^{n}\sum_{j=1}^{n}d(F_i,F_j)}{n^2} \quad (15)$$

## 4. Validation of the measures

In this section, we validate the Normalized Variance measure in terms of its ability to detect invariances in the models and compare it to the Goodfellow and ANOVA measures. We also present results that show the need for the normalization scheme in the Normalized Variance measure. Finally, we measure its sensitivity to random initializations of the models' weights, to ensure that the results of the measure do not depend overmuch on the training procedure itself and specific final weights, but on the choice of model, dataset, transformations, and other hyperparameters.

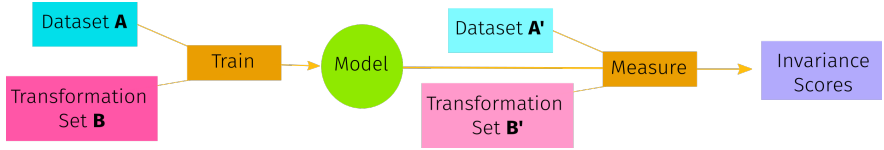

Figure 10: General diagram for our experimental methodology. We train a model using a dataset A applying data augmentation with transformation set B. The resulting model is therefore invariant to B. Afterwards, we measure the invariance of the trained model using dataset A' and transformation set B'. In most experiments, the training and measuring datasets are the same, as well as the transformations A and B.

We focus on the Normalized Variance measure because it is the most efficient. Also, given that the Normalized Distance can be seen as an approximation of the Normalized Variance for the euclidean distance, many conclusions about the Normalized Variance will be true for the Normalized Distance as well. First, we perform qualitative and quantitative experiments to determine if indeed the Normalized Variance measure is capable of measuring the desired property (invariance). Afterward, we study the general behavior of the measure in terms of its dependence on weight initialization and the dataset and transformation set sizes.

While the measures can be employed to analyze any type of model or input type, in this work we focus on image classification problems since transformations in this domain are more easily understood and defined, particularly affine transformations, using CNNs.

The general methodology of our experiments (figure 10) consists of training a model with a given dataset A and a set of transformations B used for data augmentation to force the network to acquire invariance to B during training [42, 22]. Afterward, we evaluate measures using the trained model with another dataset A' and a set of transformations B'. We note that in many cases, A=A' and B=B', so that the same datasets and transformations are used both for training and measuring.

We now describe the datasets, transformations, and models used in the experiments and analyses. All experiments can be replicated with code available at `https://github.com/facundoq/transformational_measures_experiments`.

### 4.1. Experimental setup

#### 4.1.1. Datasets

All of our experiments use MNIST and CIFAR10. Both datasets are well known and we expect any analysis performed on them is easy to understand and relate to existing methods. Also, both are small datasets to ease the computational burden. While MNIST is somewhat toy-like, it provides more interpretable results. Since all models obtain an accuracy near 100 for the test set on MNIST, this dataset allows evaluating the results of the measure in a near-perfect accuracy scenario. CIFAR10, on the other hand, consists of more complex natural images that complement the analysis, since the model achieves accuracies around 75% in most cases.

Invariance can be measured using the training or test subsets, in the same way as other metrics such as accuracy or mean squared error. For invariance, however, it is not clear that the test subset is always more appropriate to understand a model. The training set invariance shows how the model learned the invariance of the model, and the test set invariance shows how this model's invariance representation generalizes to new data from a similar distribution.

Nonetheless, we have experimented measuring invariance with the train and test subsets (not shown for brevity) and we have found that for the MNIST and CIFAR10 datasets the invariance measures yield the same results for both subsets. Therefore, we choose to measure the invariance on the standard test sets of MNIST and CIFAR10, which can provide stronger assurances on the generality of the invariance of the model.

*4.1.2. Transformations*

We chose three common transformation sets used throughout the experiments: rotations, scalings, and translations. These sets represent common affine transformations and therefore provide a wide range of diversity in order to establish properties of the measures independently of the specific transformations used. In all cases, the transformation sets include the identity transformation.

1. Rotation (25 transformations), discretized into 25 distinct angles (including 0°). Rotations are always with respect to the center of the image.
2. Scaling (25 transformations). We scaled images by a set of 8 scale factors chosen uniformly from $(0.5, 1.25)$. We generate all possible combinations of these factors so that for each factor s, we scale the image by $(1, s)$, $(s, 1)$ and $(s, s)$, where $(s_h, s_w)$ indicate the scaling factors for the height and width of the image, respectively. Thus, we modify the aspect ratio in $\frac{2}{3}$ of the transformations. Note that we scale the contents of the image but the size is kept constant. When downscaling, we fill the borders with reflections instead of filling them with a constant color to maintain the original distribution of the pixels as much as possible. The scale factors are chosen asymetrically (0.5 vs 1.25) since upscaling the image by more than 1.25 tends to remove important parts of the object.
3. Translation (25 transformations): We used 3 translation factors: 15%, 10% and 5%. For each translation factor t, we translate the images by $\left[\,(-t,-t)\ (-t,t)\ (t,-t)\ (t,t)\ (0,t)\ (t,0)\ (0,-t)\ (-t,0)\,\right]$, for a total of $8 * 3 = 24$ non-identity translation transformations.

We will refer to these simply as the rotation, scale, translation sets. Figure 11 shows examples of all sets for MNIST and CIFAR10.

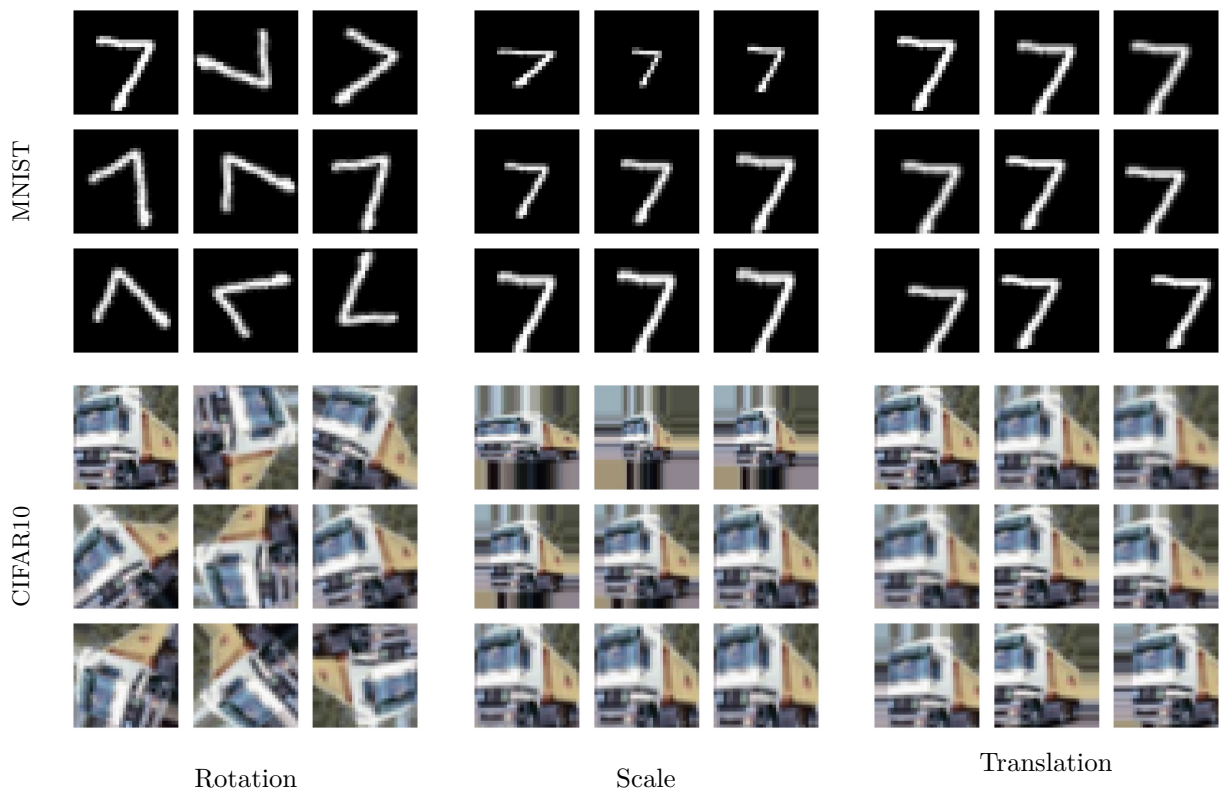

Figure 11: Samples of MNIST (top row) and CIFAR10 (bottom row) for each set of transformations.

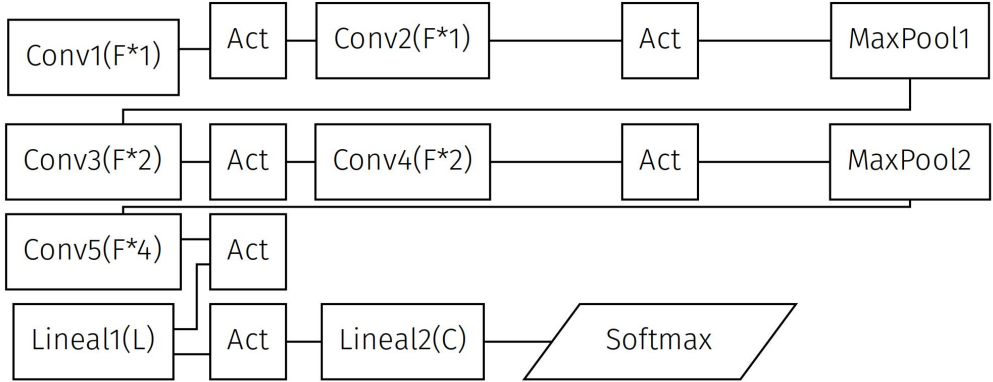

Figure 12: Architecture of the SimpleConv model. The model is a typical CNN with convolutional, max-pooling and fully connected layers.

*4.1.3. Models*

For most of the experiments we use the **SimpleConv** model, shown in figure 12. It is a simple model consisting of traditional Convolution (Conv), MaxPooling (MaxPool), and Fully Connected (FC) layers. The activation functions are all ELU, except for the final Softmax. All convolutions have stride = 1 and kernelsize = $3 \times 3$. MaxPooling layers are 2D and use stride = 2 and kernelsize = $2 \times 2$. SimpleConv was the simplest model we found with only three layers that obtains 80% accuracy in CIFAR, which is not state of the art but of similar accuracy to other more complex models. Limiting the design to only these layers applied in a feedforward fashion also facilitates the analysis.

Since our goal is not to obtain state-of-the-art accuracies, to prioritize consistency and simplicity we employed the AdamW optimizer [43] with a learning rate of $10^{-4}$ to train the models. The number of epochs used to train each model was determined separately for every dataset and set of transformations to ensure the model converges. To this effect, a base number of epochs was chosen for every model/dataset combination. To account for the difficulty of learning more transformations of the data augmentation of the training set, that number of epochs was multiplied by log(m), where m is the size of the transformation set.

Given that CIFAR10 is a more challenging dataset than MNIST, the models for MNIST have been modified to use half the number of filters/features than for CIFAR10 in all layers. Since effective invariance cannot be separated from accuracy, we verified that in all cases the accuracy of the models was superior to 95% on MNIST, and to 75% on CIFAR10.

*4.1.4. Visualization*

While comparing the full result of the measure (as a heatmap or other representation) between different models would yield the most detailed information, comparing individual activations of different trained models is very hard given that their function changes for different sets of trained weights. Since our objective is to compare measures/models and many experiments require different trained instances of the same model, we present the results of the measures aggregated by layers, which are more stable in their function. The aggregation consists of computing the mean value of the measure for all activation of each layer, and we plot the resulting means, as show in figure 14. This aggregation, therefore, allows a high-level view of the invariance of a model, which is ideal for comparing the results of different models.

*4.2. Comparison of measures*

We compare the invariance values for different measures to gauge their differences. Figure 15 shows the Normalized Variance (NV), Goodfellow (GF) and ANOVA measures on the SimpleConv model.

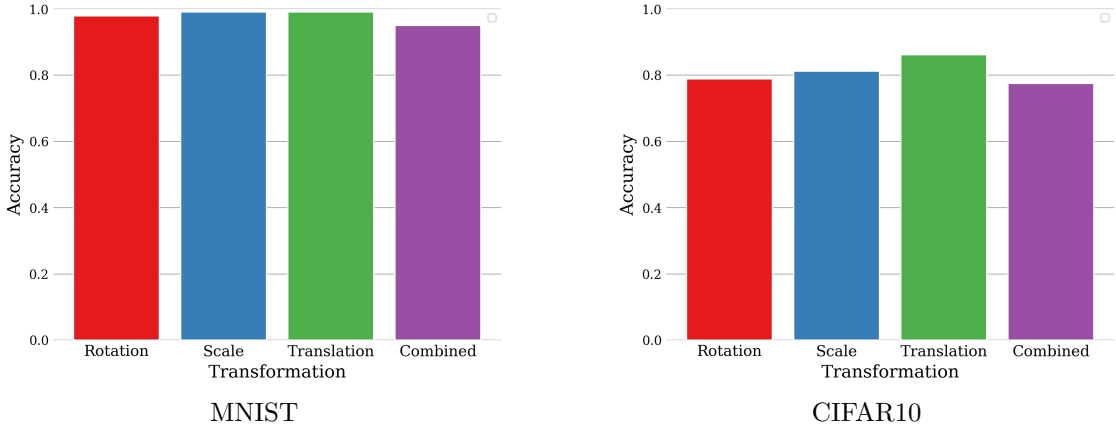

Figure 13: Accuracies for the SimpleConv model on MNIST and CIFAR10 for the 3 sets of transformations.

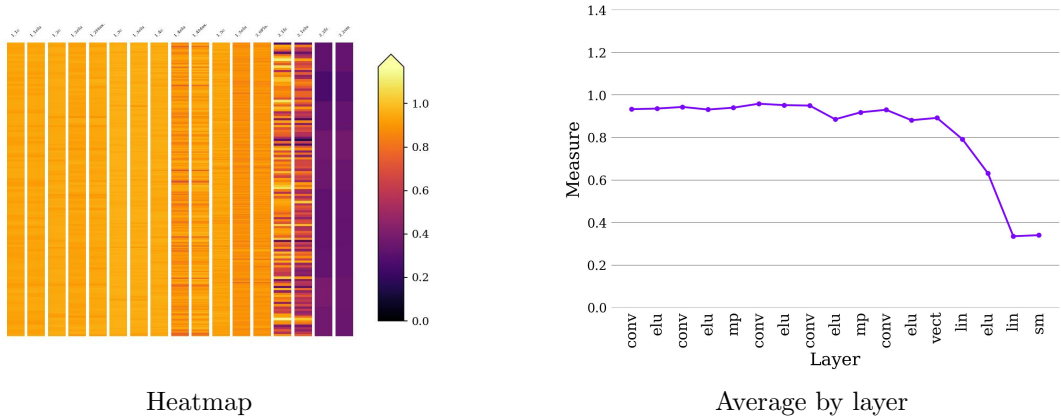

Figure 14: Values of the Normalized Variance measure for the SimpleConv model trained and measured with the MNIST dataset and rotation transformations. The results are visualized via a heatmap showing the full set of values (left) and a plot showing the mean value for each layer (right).

The ANOVA measure is insensitive to the variance of the model. It rejects the null hypothesis in almost all cases and therefore considers all activations as variant (value of 1). We note that to compute the ANOVA measure we used a Bonferroni correction with $\alpha = 0.99$ to account for the multiple hypotheses tested.

The Goodfellow measure shows less invariance for convolutional layers than for the subsequent activation function. As we show in appendix Appendix A, activation functions such as ELU never increase the variance, and therefore this suggests that the Goodfellow measure may not be a good indicator of the invariance of the model.

We note that in order to calculate the Goodfellow measure [7] in a reasonable time, we adapted the original algorithm to determine the threshold t so that instead of calculating the 1% percentile, we calculate the value z for which a normal distribution satisfies $P(f \leq z*) = 0.01$. Afterward, we employ z∗ as the threshold t in the original algorithm. In this way, we avoid having to store all activations to calculate the percentile, which would be computationally prohibiting for large models, datasets, or transformations sets. Nonetheless, the interpretation remains the same since this alternative may change the value of the threshold but not the way the measure is computed.

The ANOVA measure is very sensitive to violations of invariance and therefore not very useful for measuring that property. However, it does serve as a principled first approach to defining an invariance

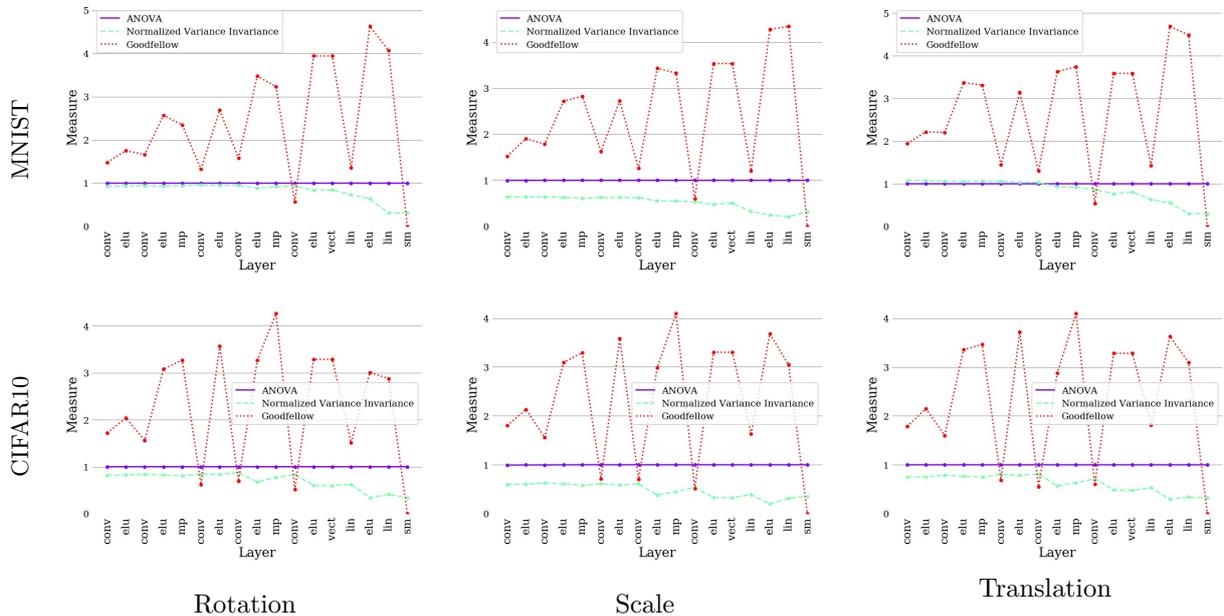

Figure 15: Comparison of normalized measures Normalized Variance (NV), Goodfellow and ANOVA for the SimpleConv model.

measure.

### 4.3. Normalization of Normalized Variance

The Transformation Variance measure by itself is a useful measure of invariance. However, its values are not normalized in any way and comparisons between models or layers can be difficult. Figure 16 shows the results of the Transformation Variance and Sample Variance measures. The models were trained with data augmentation with a set of transformations T and then the TransformationVariance and SampleVariance were also calculated with respect to T.

The magnitudes of both measures are similar for the same layer but quite different between datasets and mildly different between transformations. We note as well that the magnitudes vary significantly across layers. The units of the activations of convolutional layers are significantly lower than those of fully connected layers, hence the lower variance. Therefore, comparing the values of the Transformation Variance for convolutional and linear layers is difficult. The variance for models on MNIST is also much lower than the variance for models trained on CIFAR10. This is expected given that the background for MNIST images is much more uniform but still makes comparisons between datasets difficult. This confirms our claims in section 3 on the importance of normalizing the Transformation Variance to obtain values that are interpretable across layers, datasets, and transformations.

### 4.4. Dependence on size of dataset and transformations

We analyze the Normalized Variance measure in terms of the datasets and transformation sets used to measure it, varying their size systematically and independently. In this way, we can gain an initial understanding of the computational requirements of the measures. In this case, we consider as reference the value of the measure computed with 2304 samples and transformations. As figure 17 shows, the relative error between computing the measure with both 2304 samples and transformations ($\sim 5300000$ values in the ST matrix) is at most 10.6% with respect to computing the measure with only 24 samples and transformations ($\sim 600$ values in the ST matrix). As a middle ground, choosing 385 samples and transformations yields a relative error of at most %1.5 with only $\sim 160000$ values in the ST matrix, which seems a reasonable tradeoff. We also note that errors always decrease when increasing either the number of samples or transformations, indicating a convergence with larger sizes. This indicates that the measure is well behaved in its dependence on the size of the dataset and transformation set.

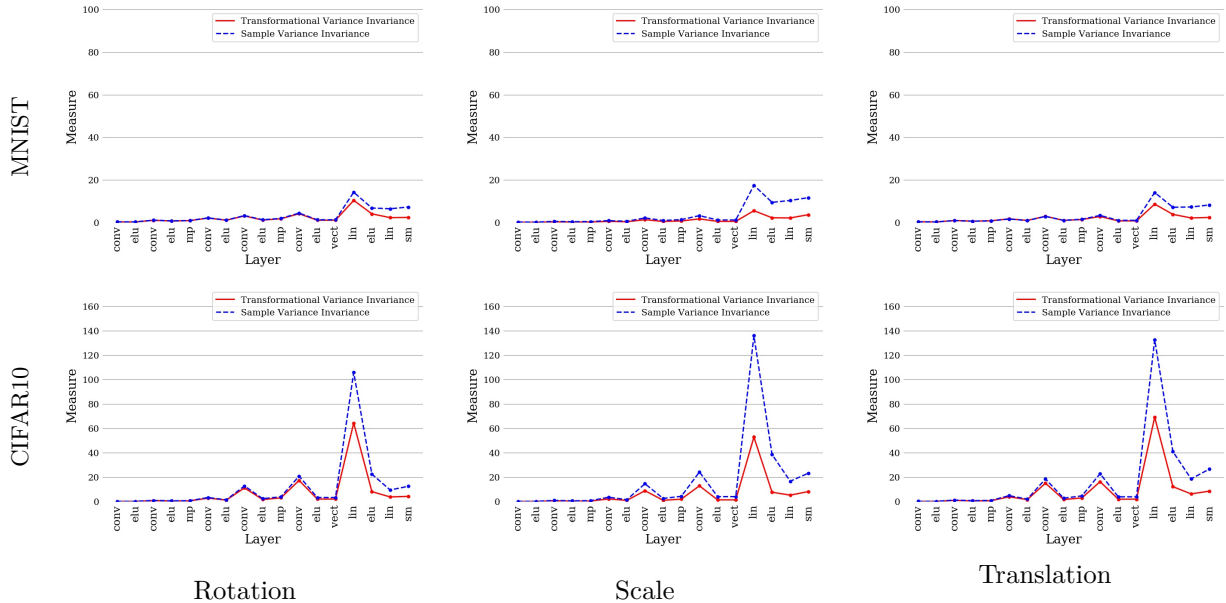

Figure 16: Comparison of unnormalized measures Transformation Variance and Sample Variance for the SimpleConv model. The scale of the values of the measure differ with the choice of dataset and transformation set.

### 4.5. Correlation between invariance and measures

While an invariance measure can be theoretically sound, it may not be useful in practice if has poor sensitivity to the changes in the model's invariance. However, there are no previously proposed methods or tests to validate the appropriateness of an invariance measure.

In order to verify the correlation of the Normalized Variance measure, we use the fact that models trained without data augmentation have very low accuracy when evaluated on transformed samples[29]. Models trained *with* data augmentation, on the other hand, mostly recover the lost performance. Therefore, we expect the latter models to possess more invariance in their activations, or at the very least in the final Softmax layer [42, 22]. In this way, we can determine if the measures are sensitive to small changes in invariance, and furthermore, if they show a positive correlation between the amount of data augmentation in training and the measured invariance.

With these assumptions, we can train different models with subsequently more complex transformations of the same type. Then, we measure the invariance of each model with respect to the most complex of these transformations. Ideally, the measure should detect the increase of invariance in the subsequent models.

Figure 18 shows the results for the different training transformations complexity for the Normalized Variance measure. The measure was always evaluated with the most complex of the transformation sets. We can observe that the Normalized Variance measure captures the increasing invariance of the model's activations. The change in invariance appears to occur mostly in the linear layers and occasionally but less significantly in the last convolutional layers.

The Goodfellow measure (figure 19), on the other hand, fails to capture this relationship, and even more, it measures *less* invariance for models trained with data augmentation in many cases.

### 4.6. Stability of the Normalized Variance measure

To study the effect of the random initialization of the model on the stability of the measure, we trained N=30 instances of each model using the same architecture, dataset, and set of transformations, until con-vergence. Each model was initialized with different random weights. Afterward, each trained model was evaluated with the Normalized Variance measure.

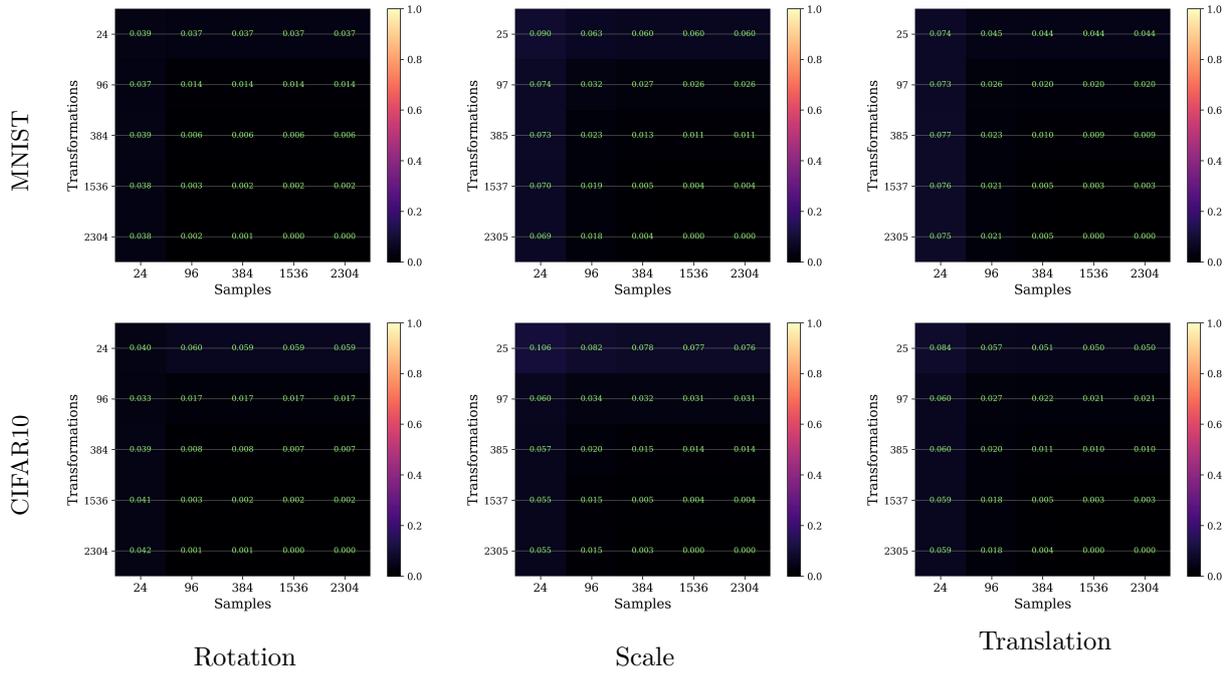

Figure 17: Relative error between the Normalized Variance of the same trained model, type of transformations and data, but varying the size of the transformation set and dataset. The relative error is computed with respect to the values of the measure computed with the largest transformation set/dataset combination.

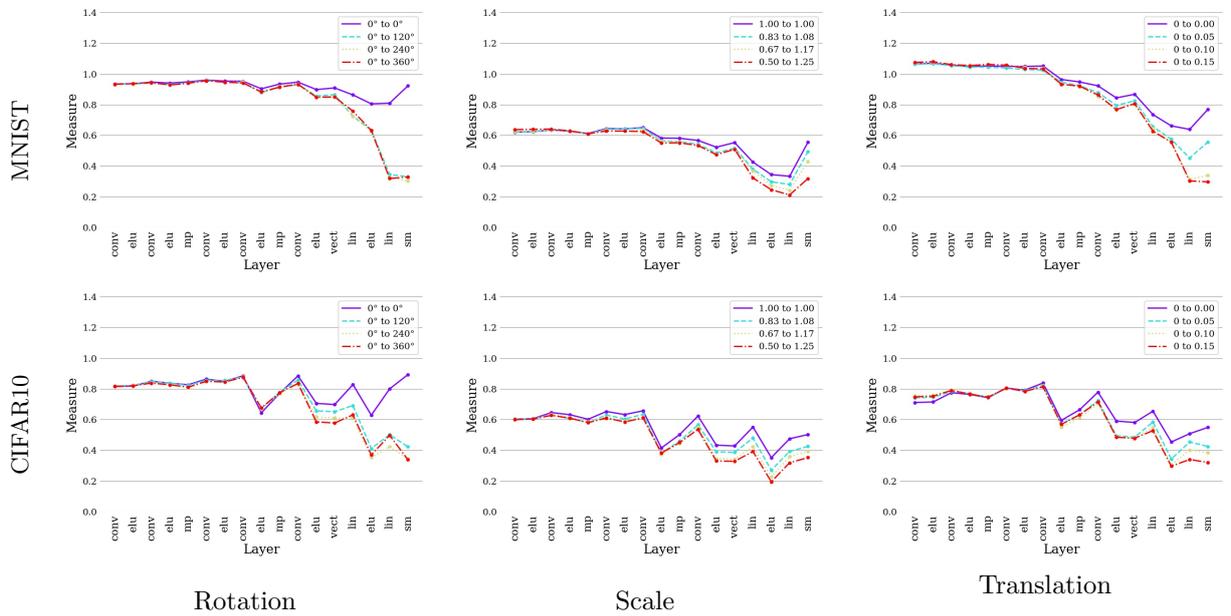

Figure 18: Normalized Variance measure of SimpleConv models trained with different sets of transformation (including a singleton set with the identity transformation, which corresponds to no data augmentation). The models were evaluated with the most complex transformation set in each case.

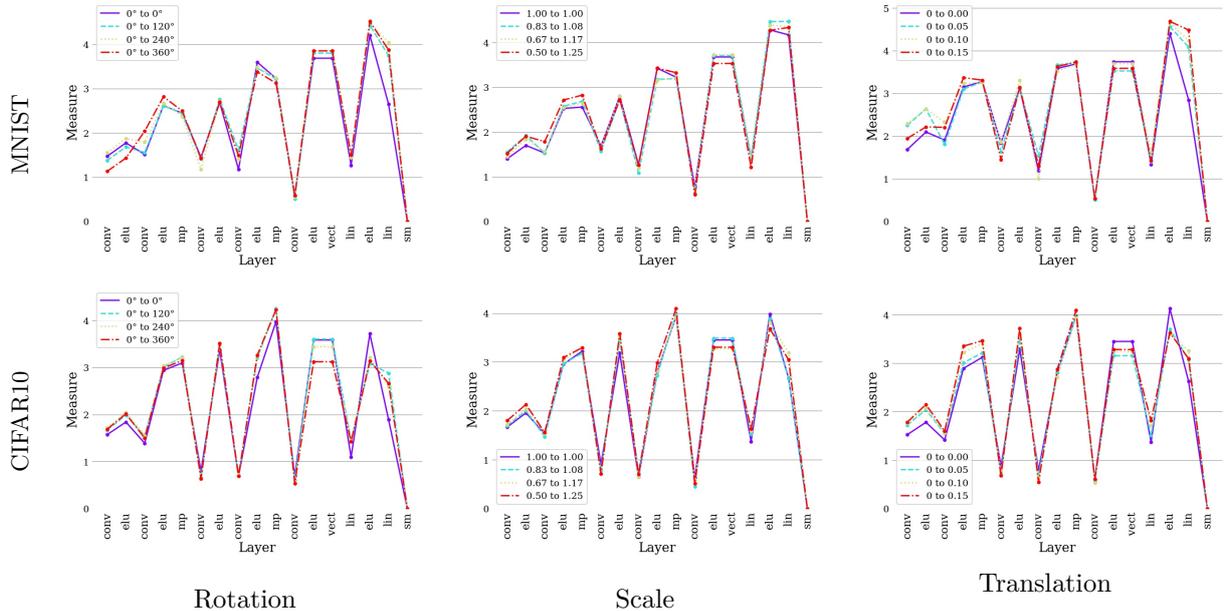

Figure 19: Goodfellow measure of SimpleConv models trained with different transformations (including a singleton set with the identity transformation, which corresponds to no data augmentation. The models were evaluated with the most complex transformation in each case.

Since comparing the representations of each model instance directly is difficult [44], we perform a N-Sample Anderson-Darling test to verify if the distribution of invariance of the N models is the same or if it differs [45]. The tests are performed by layer, so that we obtain a p-value for each layer. A higher p-value indicates that the distribution for that layer is not the same for all models. Note that p-values are expressed in terms of type-I errors, since the null hypothesis is that all N models have the same distribution of invariance. However, given the large number of models we are comparing, the chances of finding any two models with a different distribution grows as the number of models increases. Therefore, by using a large N in this case we are also substantially reducing the chance of type-II errors.

The results (figure 20) suggest that the measures vary very slightly with respect to the initialization. Only a few cases we encounter p-values larger than 0.1%. This indicates that the pattern of mean invariance per layer is an emergent property despite the random initialization of the network weights. Therefore, this pattern might mostly depend only on the model architecture, dataset, and transformation.

In any case, the stability of the measures with respect to the initialization is a desirable property, given that it is not always computationally possible to repeatedly train models to achieve a greater degree of certainty of the invariance of a model.

Also, note that the pattern of invariance does vary when changing either the transformation or dataset. This indicates that different transformations may require different ways of encoding the invariance, and also that the invariance to a set of transformations is actually specialized for the a specific dataset. This has consequences in terms of the transferability of the representations.

*Random Weights.* While evaluating the measure with different trained models indicates if it is stable for different initializations, it is also interesting to understand the base invariance of an untrained model and if it varies significantly for different initializations. Therefore, we measured the Normalized Variance of N = 30 different initializations of the same model. As figure 21 shows, similarly to the trained case, the values of the measure for different untrained weights does not differ significantly. This reinforces the view that the distribution of invariance of each layer is dictated mostly by the model, dataset, and transformation set, and not the specific set of weights of the model.

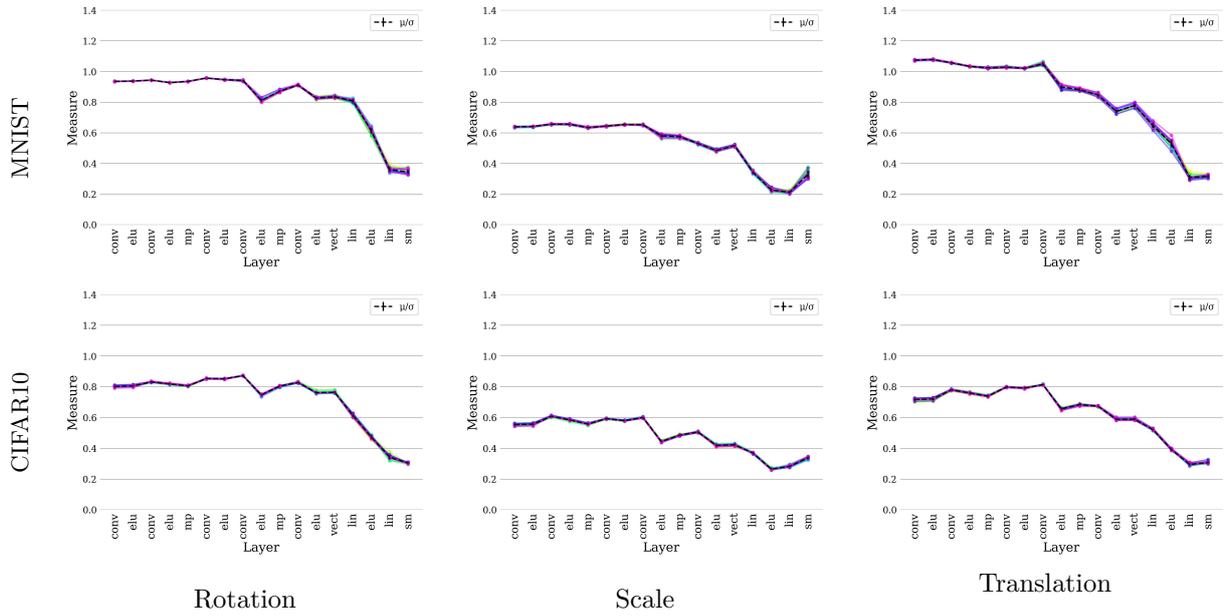

Figure 20: Stability of the Normalized Variance measure for the SimpleConv model. Each color line in each plot corresponds to the invariance obtained by different models with the same architecture. The dashed line black shows the mean values. Vertical black bars indicate standard deviation (very low in most cases). Additionally,

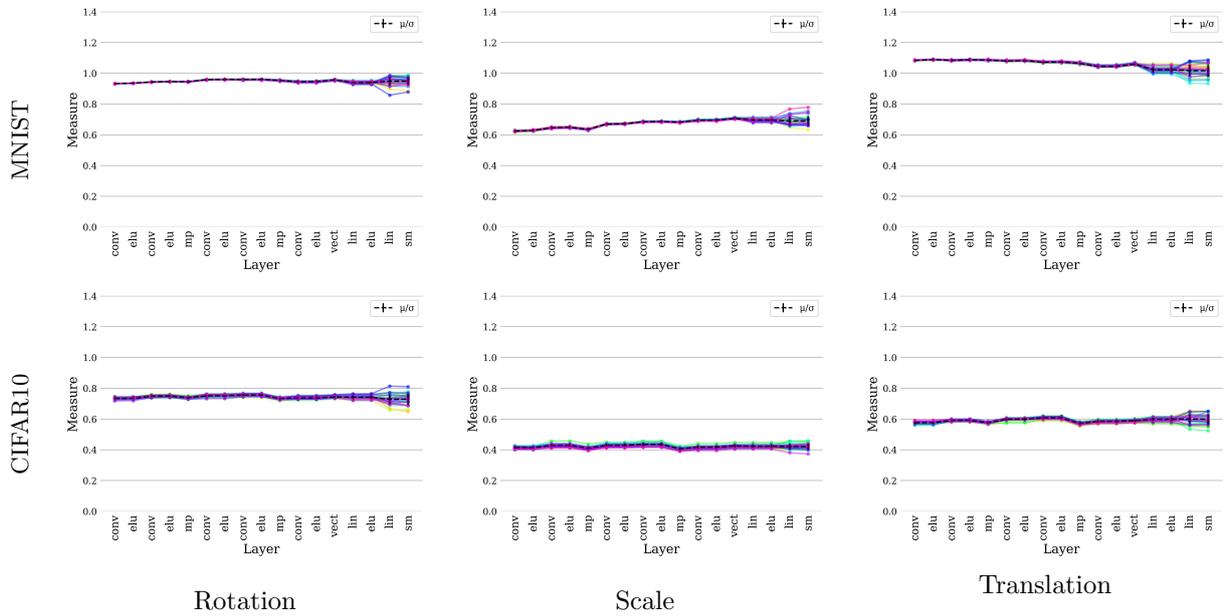

Figure 21: Normalized Variance measure for *untrained* models with different initializations. Each color line in each plot corresponds to the invariance obtained by different models with the same architecture. The dashed line black shows the mean values. Vertical black bars indicate standard deviation.

*Invariance while training.* To complement this view, we can attempt to understand the dynamics of *learning* invariances. Therefore, we computed invariance measures at regular intervals while the models were being trained. The regular intervals were chosen as percentages of the total number of epochs. We use the percentages 0%, 1%, 3%, 5%, 10%, 30%, 50%, 100% for all datasets/transformations.

Figure 22 shows the distribution of the invariance over the layers of a model while it is being trained. In all cases, we can observe that after the first one or two epochs, the invariance structure of the network remains mostly fixed. Also, this structure is different from the structure of the networks at epoch 0, that is, with random, untrained weights.

The variance of the lower convolutional layers changes only slightly throughout the training. That of the final convolutional layers and the FC layers does change more significantly during training, mostly becoming smaller. In the case of MNIST, this transition is much weaker faster for CIFAR10, possibly because of the different complexities of the dataset.

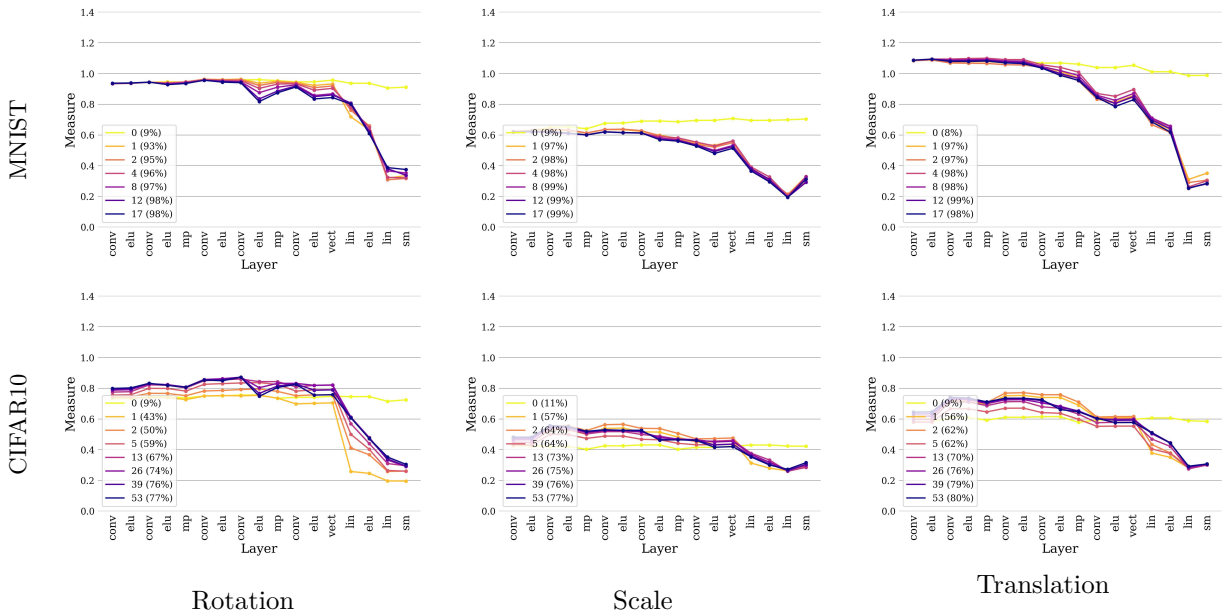

Rotation  Scale  Translation

Figure 22: Comparison of the Normalized Variance measures for the SimpleConv model. Each line in each plot corresponds to the same model in a different epoch. Percentages indicate the corresponding test set accuracy in that epoch. For MNIST, train percentages 0% and 1% were mapped to epoch 0, and 5% and 10% to epoch 1, because of the low number of epochs.

Recalling the stability of the measure for the final values of the weights, the results indicate that given a fixed dataset, transformation, and model, the distribution of invariance before and after training is quite similar. Therefore, training with data augmentation forces, through the loss function, models to converge to a similar invariant representation despite the initial random initialization.

## 5. Conclusions and future work

Invariance can be a useful or necessary property for a network in different components of its architecture. However, there are several ways to achieve and encode invariance in a network, and there is a need to understand their differences. This work proposes invariance measures that enable important tools to investigate these encodings. The measures can be applied to any neural network model, dataset and finite transformation set.

Along with the measures, we propose experimental methods that take advantage of the network to analyze non-trivial properties of the networks and their encoding. In this case, our measure can quantify the invariance in a simpler way that is more interpretable and sensitive than previous approaches.

Our main findings show that the measures are indeed able to effectively quantify the invariance of a network. Also, we have shown that the measures are efficient in terms of sample and transformation set sizes, and stable with respect to the random initialization of network weights.

Furthermore, our first analysis of a simple CNN model, made invariant by data augmentation, revealed interesting facts about this architecture. For instance, the structure of the invariance of the layers converges to the same distribution, even when the network's weights are initialized randomly. Also, the invariance structure of randomly initialized networks without training is very similar. This indicates that the encoding of invariance achieved by data augmentation can be stable and reliable.

In future work, we expect to expand the set of measures following a similar approach to also quantify same-equivariance and equivariance. Another limitation of the presented measures is that they focus on individual activations to compute the measures, and aggregate these results to perform global analysis in a simple fashion. It would be interesting to extend these measures to allow them to automatically detect the transformational structures of the network in terms of groups of activations. This would allow the analysis with intermediate granularities that lie between analyzing a whole network, layer, or individual activations.

Finally, while the measures apply to any type of neural networks, transformations and datasets, in this work we restricted ourselves to well-known instances of these objects, such as CNNs, affine transformations, and the MNIST and CIFAR datasets. Therefore, we are also interested in widening the set of models, transformations, and domains in which to apply the measure, ranging outside the set of affine transformations and image classification problems.

## Acknowledgments

The Titan X Pascal used for this research was donated by the NVIDIA Corporation.

# Appendix A. Variance of common activation functions

In order to characterize the evolution of invariance, it is useful to understand how activation functions can affect the variance of the activations. Many theoretical analyses of the variance of activation functions have been performed in terms of the output of a preceding fully connected or convolutional network, or the entire network[46]. However, these assume a certain distribution of the input, typically defined by the type of the previous layer or the initial input to the network.

Since our measures are agnostic to the type of layers used before an activation function, it would be useful to obtain a characterization of the variance of the activation functions irrespective of the distribution of its input. Therefore, we study common activation functions in terms of their effect on the variance of their output.

The mean E and variance V of a continuous function f of a real random variable x can be defined as:

$$\begin{aligned} E(f(x)) &= \int_{-\infty}^{\infty} f(x)p(x)dx \\ V(f(x)) &= \int_{-\infty}^{\infty} (E(f(x)) - f(x))^2 p(x)dx \\ &= E(f(x)^2) - E(f(x))^2 \end{aligned} \quad (A.1)$$

We show that $V(f(x)) \leq V(x)$ for various common activation functions $f(x)$. That is, the activation functions are non-increasing with respect to the variance of its input. In general, the cases where this inequality is strict (ie, $V(f(x)) < V(x)$) depend on $p(x)$ and therefore cannot be specified without characterizing the output of the previous layer, which would defeat the original purpose. Nonetheless, the results on section 4 provide empirical evidence that the variance is lower, on average, for the activations corresponding to activation functions than those of the immediately previous layer. In practice, therefore, our result implies that these activation functions tend to lower the variance of a network.

To show this inequality for the functions ReLU, LeakyReLU, PReLU and ELU[47], let's cast these functions into the form defined in equation [A.2], where g is an arbitrary continuous function that satisfies $x \leq g(x)$ for $x < 0$:

$$f(x) = \begin{cases} g(x) & x < 0 \\ x & x \geq 0 \end{cases} \quad (A.2)$$

The aforementioned activation functions can all be defined in terms of equation [A.2] with appropriate choices for $g(x)$. Also, they all satisfy $x \leq g(x)$ for $x < 0$, with reasonable choices for the hyperparameter $\alpha$ in LeakyReLU and ELU ($\alpha < 1$) and restrictions for the corresponding *parameter* $\alpha$ in PReLU ($\alpha < 1$). These are, coincidentally, the most common choices for these parameters.

As show below, A.3 proves that $V(f(x)) \leq V(x)$ with the help of the inequalities $E(f(x)) \geq E(x)$ and $E(f(x)^2) \leq E(x^2)$ (equations [A.4] and [A.5]).

$$\begin{aligned} V(f(x)) &= E(f(x)^2) - E(f(x))^2 \\ &\leq E(x^2) - E(f(x))^2 \\ &\leq E(x^2) - E(x)^2 \\ &= V(x) \end{aligned} \quad (A.3)$$

$$\begin{aligned}
E(f(x)) &= \int_{-\infty}^{\infty} f(x)p(x) \\
&= \int_{-\infty}^{0} g(x)p(x)dx + \int_{0}^{\infty} xp(x)dx \qquad [x \leq g(x)] \\
&\geq \int_{-\infty}^{0} xp(x)dx \quad + \int_{0}^{\infty} xp(x)dx \\
&= \int_{0}^{\infty} xp(x)dx \quad = E(x)
\end{aligned} \qquad (A.4)$$

$$\begin{aligned}
E(f(x)^2) &= \int_{-\infty}^{0} g(x)^2 p(x)dx + \int_{0}^{\infty} x^2 p(x)dx \\
&\leq \int_{-\infty}^{0} x^2 p(x)dx + \int_{0}^{\infty} x^2 p(x)dx \qquad [x \leq g(x)] \\
&= \int_{-\infty}^{\infty} x^2 p(x)dx = E(x^2)
\end{aligned} \qquad (A.5)$$